\documentclass[10pt,twocolumn,letterpaper]{article}

\usepackage{cvpr}
\usepackage{times}
\usepackage{epsfig}
\usepackage{graphicx}
\usepackage{amsmath}
\usepackage{amssymb}
\usepackage{mathrsfs}
\usepackage{multirow}
\usepackage{array}
\usepackage{subfigure} 
\usepackage{algorithms}
\usepackage{algpseudocode}  
\usepackage{algorithmicx}
\renewcommand{\algorithmicrequire}{\textbf{Input:}}  
\renewcommand{\algorithmicensure}{\textbf{Output:}} 

\usepackage[pagebackref=true,breaklinks=true,letterpaper=true,colorlinks,bookmarks=false]{hyperref}

\cvprfinalcopy 

\def\cvprPaperID{6823} 

\ifcvprfinal
\begin{document}

\title{Cooling-Shrinking Attack: Blinding the Tracker with Imperceptible Noises}

\author{Bin Yan$^{1}$, Dong Wang$^{1}\thanks{Corresponding Author: Dr. Dong Wang, wdice@dlut.edu.cn}$,  Huchuan Lu$^{1,2}$ and Xiaoyun Yang$^3$\\
$^1$School of Information and Communication Engineering, Dalian University of Technology, China\\
$^2$Peng Cheng Laboratory\\
$^3$China Science IntelliCloud Technology Co., Ltd., \\
{\tt\footnotesize yan\_bin@mail.dlut.edu.cn, 
\{wdice, lhchuan\}@dlut.edu.cn, xiaoyun.yang@intellicloud.ai}
}

\maketitle
\thispagestyle{empty}

\begin{abstract}
   Adversarial attack of CNN aims at deceiving models to misbehave by adding imperceptible 
   perturbations to images. This feature facilitates to understand neural networks deeply and to 
   improve the robustness of deep learning models.
   Although several works have focused on attacking image classifiers and object detectors, 
   an effective and efficient method for attacking single object trackers of any target in a 
   model-free way remains lacking.
   In this paper, a cooling-shrinking attack method is proposed to deceive state-of-the-art 
   SiameseRPN-based trackers. 
   An effective and efficient perturbation generator is trained with a carefully designed adversarial 
   loss, which can simultaneously cool hot regions where the target exists on the heatmaps and force 
   the predicted bounding box to shrink, making the tracked target invisible to trackers.
   Numerous experiments on OTB100, VOT2018, and LaSOT datasets show that our method can 
   effectively fool the state-of-the-art SiameseRPN++ tracker by adding small perturbations to 
   the template or the search regions. 
   Besides, our method has good transferability and is able to deceive other top-performance 
   trackers such as DaSiamRPN, DaSiamRPN-UpdateNet, and DiMP. 
   The source codes are available at 
   \href{https://github.com/MasterBin-IIAU/CSA}{https://github.com/MasterBin-IIAU/CSA}. 
\end{abstract}

\vspace{-3mm}
\section{Introduction}
Online single object tracking is a fundamental task in computer vision and has many important 
applications including intelligent surveillance, autonomous driving, human-machine interaction, 
to name a few. 
In recent years, 
as deep learning matures and large-scale tracking datasets~\cite{DVT-Review,LaSOT,GOT10K} are introduced, 
the single object tracking field has developed rapidly. 
Current state-of-the-art trackers~\cite{SiameseRPN,DSiam,SiamRPNplusplus,fan2017robust,li2017object,ATOM,DiMP} can be grouped into two categories: 
deep discriminative trackers~\cite{ATOM,DiMP,LTMU-CVPR2020}, and SiameseRPN-based 
trackers~\cite{SiameseRPN,DSiam,Deeper-wider-SiamRPN,SiamRPNplusplus}.
Deep discriminative trackers decompose tracking into two sub-problems: classification and 
state estimation. 
The first one is solved by an online-learning classifier and the second one is achieved by 
maximizing the overlap between candidates and the ground truth.
SiameseRPN-based trackers formulate tracking as a one-shot detection problem, locating 
objects that have similar appearance with the initial template on the search region in each frame. 
Considering their balance between accuracy and speed, SiameseRPN-series trackers have attracted 
more attention than deep discriminative trackers.

Adversarial attack is originated from~\cite{Intriguing-properties-NN}, which has shown that 
state-of-the-art deep learning models can be fooled by adding small perturbations to original images. 
Research on adversarial attack is beneficial to understand deep neural networks and design robust models. 
Popular adversarial attack methods can be roughly summarized into two categories: 
iterative-optimization-based and deep-network-based attacks.  
The former method~\cite{FGSM,Deepfool,DAG} applies many times of gradient ascent to maximize 
an adversarial objective function for deceiving deep networks and is usually time-consuming. 
However, the latter one~\cite{advGAN,UEA} applies tremendous data to train an adversarial 
perturbation-generator. 
The latter method is faster than the former method because only one-time forward 
propagation is needed for each attack after training.
In recent years, adversarial attack has become a popular topic and has extended from 
image classification to more challenging tasks, such as object detection~\cite{DAG,UEA} and semantic segmentation~\cite{DAG}.

However, an effective and efficient adversarial attack method for single object tracking remains lacking.
In this study, we choose the state-of-the-art SiamRPN++~\cite{SiamRPNplusplus} tracker as our main 
research object and propose a novel cooling-shrinking attack method. 
This method learns an efficient perturbation generator to make the tracker fail by simultaneously 
cooling down hot regions where the target exists on the heatmaps and forcing the predicted bounding 
box to shrink during online tracking. 
Our main contribution can be summarized as follows. 

 \vspace{-3mm}
\begin{itemize}
\setlength{\itemsep}{0pt}
\setlength{\parsep}{0pt}
\setlength{\parskip}{0pt}
\item \emph{A novel and efficient cooling-shrinking attack method is proposed to effectively fool 
the SiamRPN++ tracker. 
Experiments on OTB100~\cite{OTB2015}, VOT2018~\cite{VOT2018report}, and 
LaSOT~\cite{LaSOT} show that our method can successfully deceive the state-of-the-art 
SiamRPN++ tracker.}
\item \emph{Numerous experiments show that a discriminator is unnecessary in this task 
because existing $L_2$ loss and fooling loss have already achieved our goal.} 
\item \emph{Our attacking method has good transferability. 
Experimental results demonstrate that state-of-the-art trackers (such as DaSiamRPN and DiMP) 
can also be deceived by our method, even though this method is not specially designed for them.}
\end{itemize}

\vspace{-4mm}
\section{Related Works}

\subsection{Single Object Tracking}
Given the tracked target in the first frame, single object tracking (SOT) aims at capturing the location of 
the target in the subsequent frames. 
Different from object detection that recognizes objects of predefined categories, the SOT  
task belongs to one-shot learning, requiring trackers to be capable of tracking any possible targets. 
Efficient and robust trackers are difficult to be designed because of challenges, such as occlusion, 
similar distractors, deformation, and motion blur, during tracking.
Recently, with the prosperity of deep learning and the introduction of large-scale object tracking 
datasets~\cite{LaSOT,GOT10K}, the study of SOT has undergone a rapid development. 
Currently, state-of-the-art trackers can be divided into two categories. 
One is based on SiamRPN (including SiamRPN~\cite{SiameseRPN}, DaSiamRPN~\cite{DSiam}, 
SiamRPN+~\cite{Deeper-wider-SiamRPN}, SiamRPN++~\cite{SiamRPNplusplus}, and 
SiamMask~\cite{SiamMask}), and the other is based on deep discriminative models
(including ATOM~\cite{ATOM} and DiMP~\cite{DiMP}).

SiamRPN~\cite{SiameseRPN} formulates SOT as a one-shot detection problem and 
is the first attempt to introduce RPN~\cite{FasterRCNN} in the tracking filed. 
With the help of RPN, SiamRPN removes heavy multi-scale correlation operations, 
running at a high speed and producing accurate results. 
DaSiamRPN~\cite{DSiam} relieves the SiamRPN's weakness of being susceptible to distractors, 
by introducing challenging samples to the training set. 
However, the negative effect of image padding makes SiamRPN and DaSiamRPN only apply the 
shallow and padding-free AlexNet's variant as their backbone, which does not fully take advantage 
of the capability of modern deep neural networks~\cite{inception,ResNet}. 
To overcome this problem, some studies have proposed the addition of a cropping-inside residual 
unit and a spatial-aware sampling strategy in SiamRPN+~\cite{Deeper-wider-SiamRPN} and 
SiamRPN++~\cite{SiamRPNplusplus}. 
These works relieve the center bias problem caused by image padding, making the SiameseRPN 
framework benefit from modern backbones and significantly improving the Siamese tracker's 
performance.
SiamMask~\cite{SiamMask} proposes a unified framework for visual object tracking and 
semi-supervised video object segmentation, further increasing the accuracy of predicted bounding boxes.

ATOM~\cite{ATOM} proposes a tracking framework composed of the dedicated target 
estimation and classification components. 
The target estimation module is an IOU-Net's variant~\cite{IOU-Net} that can produce an accurate 
bounding box of the target, given the initial appearance information. 
DiMP~\cite{DiMP} inherits the ATOM's framework (making it end-to-end trainable) and proposes 
a more discriminative model predictor. 
DiMP achieves state-of-the-art performance on most tracking benchmarks, thereby serving a strong baseline 
in the tracking community.

\subsection{Adversarial Attack}
The adversarial attack in~\cite{Intriguing-properties-NN} indicates that CNN is 
highly vulnerable to attack and state-of-the-art classifiers can be easily fooled by adding 
visually imperceptible noises to original images.
Since then, several works have focused on adversarial attacks.
Early works~\cite{FGSM,Deepfool,DAG,advGAN,UEA} add perturbations in the digital world, 
directly changing pixel values which are fed into the networks. 
Latter works focus on creating physical adversarial objects, such as eyeglasses~\cite{advface}, 
posters~\cite{advtexture}, and animals~\cite{EOT}, in the real world, further broadening 
the influence of adversarial attacks.

Digitally adversarial attack methods can be roughly divided into two categories: iterative-optimization-based 
and deep-network-based algorithms. 
The former ones (including FGSM~\cite{FGSM}, Deepfool~\cite{Deepfool}, and DAG~\cite{DAG}) optimize
an adversarial objective function to fool deep networks and are usually time-consuming due to 
several iterations. 
However, deep-network-based methods (including advGAN~\cite{advGAN} and UEA~\cite{UEA}) use 
tremendous data to train a generator for adding perturbations. 
The latter type is generally faster than the former type because the operation for transforming an image to 
an adversarial one does not need to be repeated.
In specific, FGSM~\cite{FGSM} hypothesizes that neural networks behave in very linear ways and 
proposes a ``Fast Gradient Sign Method" to attack them. 
Deepfool~\cite{Deepfool} generates adversarial examples by pushing data points around the classification 
boundary past it with minimal perturbations. 
AdvGAN~\cite{advGAN} is the first work to generate adversarial examples with GAN, which can run 
efficiently during the inference phase.
Recent research on adversarial attack has extended from image classification to more challenging tasks, 
such as object detection. 
Two impressive works are iterative-optimization-based DAG~\cite{DAG} and deep-model-based UEA~\cite{UEA}. 
DAG~\cite{DAG} sets the difference of the classification score between adversarial and 
ground-truth classes as its objective function, and then optimizes it using gradient ascent. 
Although it achieves a high fooling rate, it is time-consuming because several iterations are needed. 
UEA~\cite{UEA} chooses GAN as a core component to generate perturbations and trains the generator 
with a carefully designed adversarial loss. 
UEA achieves comparative performance but is much faster than DAG, because only one-time forward 
propagation is needed during the inference phase after training.

Adversarial objects in the physical world are more difficult to generate than digital adversarial examples 
because the literature~\cite{adv-autonomous,alleviate-adv} has revealed that adversarial examples 
generated by standard methods are not robust to the common phenomena in the physical world, 
such as varying viewpoints and camera noises. 
To overcome this problem, the expectation over transformation (EOT)~\cite{EOT} method 
requires not only the original single example but also its augmented examples to be confusing. 
Combining with the 3D-printing technique, EOT can synthesize robust physical adversarial objects. 
Inheriting similar ideas, the method in~\cite{advface} generates adversarial eyeglasses 
that can fool state-of-the-art face recognition systems, and the method in~\cite{advtexture} 
creates an inconspicuous poster that can deceive the simple regression-based tracker GOTURN~\cite{GOTURN}.

\vspace{-3mm}
\section{Cooling-Shrinking Attack}
In this work, we propose an adversarial perturbation-generator for deceiving the SiamRPN++ tracker. 
The goal of our method is to make the target invisible to trackers, thereby leading to tracking drift. 
To accomplish this goal, we train the generator with a carefully designed and novel cooling-shrinking loss.
Considering that SiamRPN-based trackers~\cite{SiameseRPN,DSiam,SiamRPNplusplus,Deeper-wider-SiamRPN,SiamMask} 
locate the target in a local search region based on the template given in the initial frame, we design two versions of 
perturbation-generators to attack the search regions and the template respectively.
\vspace{-3mm}
\subsection{Overview of SiamRPN++}
\begin{figure}[!h]
\begin{center}
\includegraphics[width=1.0\linewidth,height=0.55\linewidth]{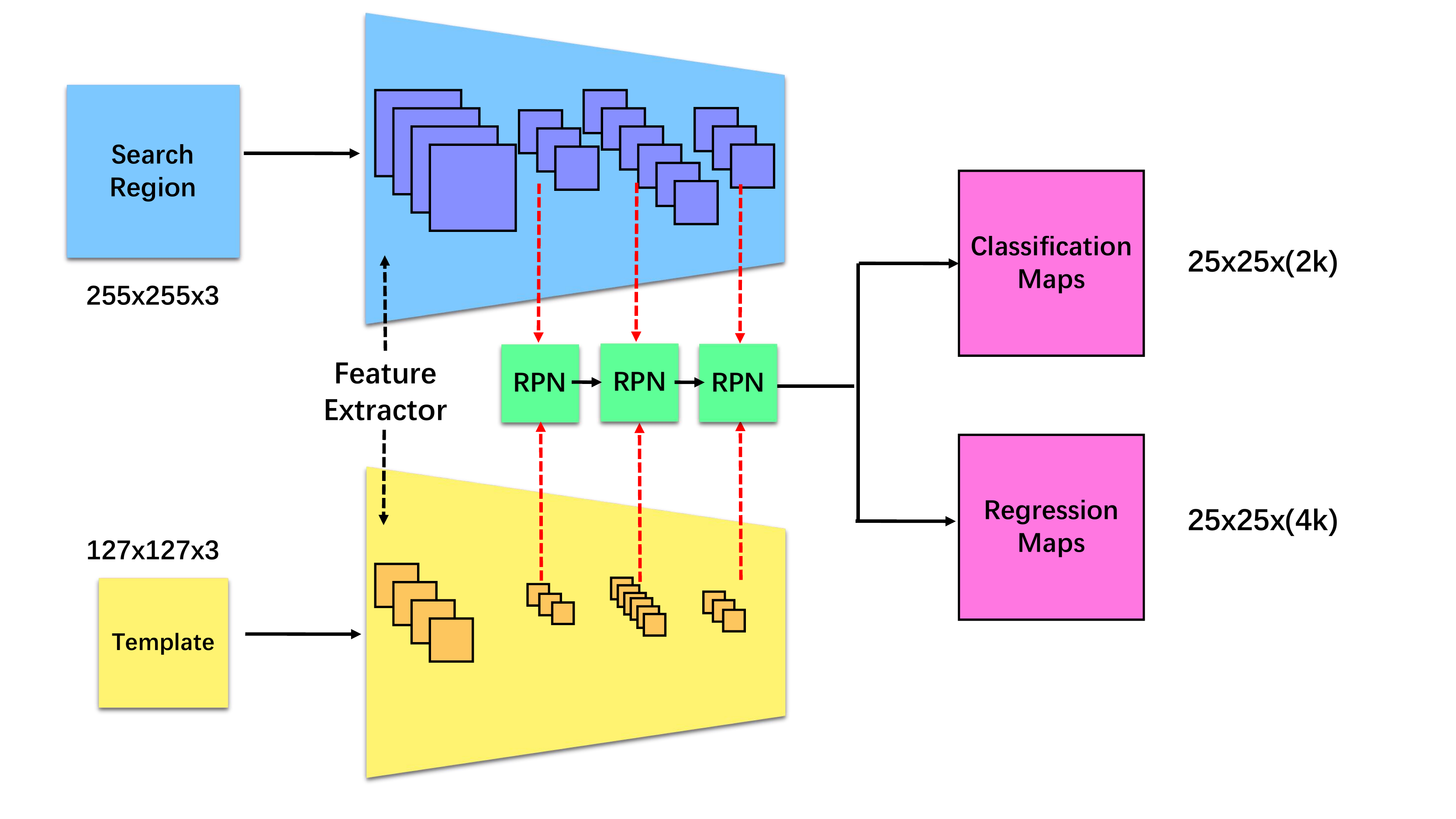}
\end{center}
\vspace{-3mm}
\caption{Network architecture of SiamRPN++~\cite{SiamRPNplusplus}. Better viewed in color with zoom-in.}
\vspace{-5mm}
\label{fig-siamRPN}
\end{figure}

So far, SiamRPN++~\cite{SiamRPNplusplus} is the most powerful SiameseRPN-based tracker, achieving 
state-of-the-art performance on almost all tracking datasets. 
The network architecture of SiamRPN++ is shown in Figure~\ref{fig-siamRPN}. 
Given the template $\mathcal{T}$ in the initial frame, SiamRPN++ detects target in the search region $\mathcal{SR}$ from the current frame. 

To be specific, the template is an image patch cropped in the first frame, providing the target's appearance 
information for the  tracker. 
The tracked target generally does not move too much between two adjacent frames.  
Most modern trackers only locate the target in a small search region centered in the position of the previous frame, 
rather than the whole image. 
The size of the search region in the current frame is proportional to the size of the target in the previous frame. 
In each frame, the template $\mathcal{T}$ and search region $\mathcal{SR}$ are first passed through a shared 
backbone network such as ResNet50~\cite{ResNet}, and their features are processed by some non-shared 
neck layers and fused by depthwise correlation. 
Based on these features, the RPN head layers predict the classification maps $\mathcal{M_C}$ and regression 
maps $\mathcal{M_R}$. 
Specifically, SiamRPN++ produces four regression factors, two of which are related to 
the center offset and the other two are responsible for the scale changes. 
Finally, the tracker considers the position with the highest classification score as the optimal target location, 
and then uses the corresponding regression factors to obtain an accurate bounding box as a result of 
the current frame.
Thus, if the final classification maps $\mathcal{M_C}$ or regression maps $\mathcal{M_R}$ are interfered, 
the tracker may lose the ability to locate the target or produce inaccurate results, leading to tracking failure.

\subsection{Overall Pipeline}
\begin{figure}[!h]
\begin{center}
\includegraphics[width=1.0\linewidth,height=0.6\linewidth]{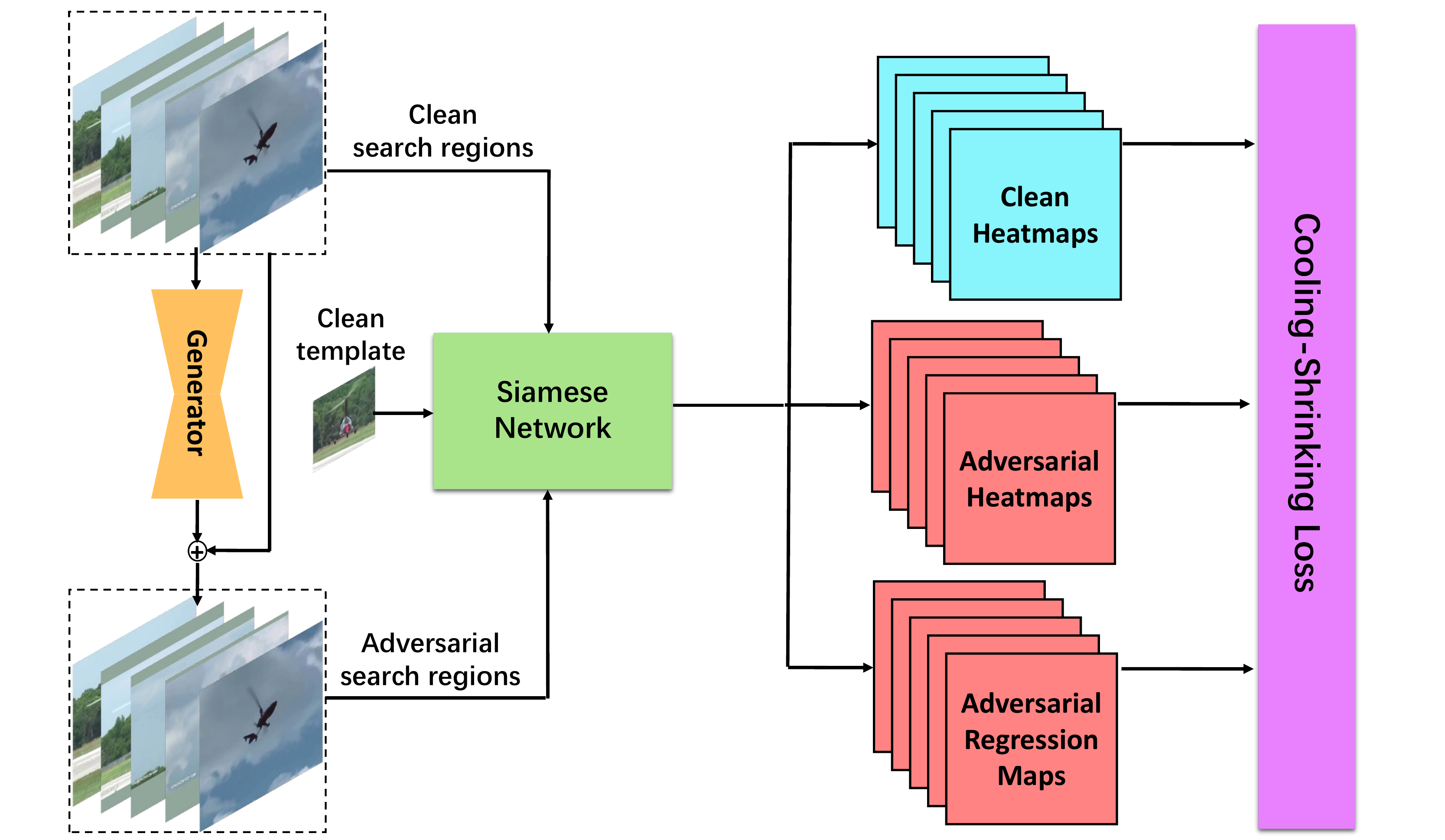}
\end{center}
\vspace{-5mm}
\caption{Network architecture of perturbation-generator for search regions. Better viewed in color with zoom-in.}
\label{fig-search}
\end{figure}

Since the pipelines of attacking the template and attacking the search regions are quite similar, we only 
discuss attacking search regions for simplicity. 
The overall pipeline of the training generator for attacking search regions is shown in Figure~\ref{fig-search}. 
During the training process, we first feed $N$ pre-cropped 
unperturbed search regions into the perturbation-generator, adding imperceptible noises to them. 
Together with a clean template, these perturbed search regions are fed into the SiamRPN++ tracker's 
Siamese network, which produces adversarial classification and regression maps of the corresponding 
search regions. 
The SiamRPN++ tracker considers regions with the highest response on the classification maps 
(heatmaps) as the target. 
Thus, regions, where the tracked target exists on the adversarial heatmaps, are expected to 
have low response values.
To indicate these regions, we also feed originally unperturbed search regions into the Siamese network, 
producing clean heatmaps. 
Then, an adversarial cooling-shrinking loss and an $L_2$ loss are applied together to train our perturbation-generator. 
The detailed training algorithm of the generator for attacking search regions is shown in 
Algorithm~\ref{alg::attackSR}.
During the online-tracking phase, to deceive the tracker, we only need to pass a clean search region into the generator, 
obtaining a new adversarial one in each frame.
\begin{algorithm}[h]
    \caption{Framework of training perturbation-generator to attack search regions}
    \label{alg::attackSR}
    \begin{algorithmic}[1]
      \Require
        $R^c$: clean search regions;
        $T^c$: clean template;
        $g_0$: randomly initialized generator
      \Ensure
        trained generator $g^{*}$
      \State Initialize generator $g_0$. Initialize siamese model $S$ and freeze its parameters;
      \Repeat
        \State Get a clean template $T^c$ and a batch of clean search regions $R^c$;
        \State Feed $T^c$ into $S$;
        \State Generate adversarial noises $P=g(R^c)$ for $R^c$;
        \State Get adversarial search regions $R^a=R^c+P$;
        \State Get adversarial heatmaps and regression maps $M_H^a,M_R^a=S(R^a,T^c)$ using $R^a$;
        \State Get clean heatmaps $M_H^c=S(R^c,T^c)$;
        \State Compute cooling loss $L_C$ and shrinking loss $L_S$ based on $M_H^a$, $M_R^a$, $M_H^c$;
        \State Compute $L_2 $ loss $L_2=\frac{1}{N}||{R^a}-{R^c}||_2$; 
        \State Compute total loss $L={\alpha_1}L_C+{\alpha_2}L_S+{\alpha_3}L_2$;
        \State Compute gradient of $L$ to generator $g$'s parameters and update with the Adam optimizer.
        \Until{model converges}
    \end{algorithmic}
  \end{algorithm}

\vspace{-3mm}
\subsection{Cooling-Shrinking Loss}
\begin{figure}[!h]
\begin{center}
\includegraphics[width=1.0\linewidth,height=0.60\linewidth]{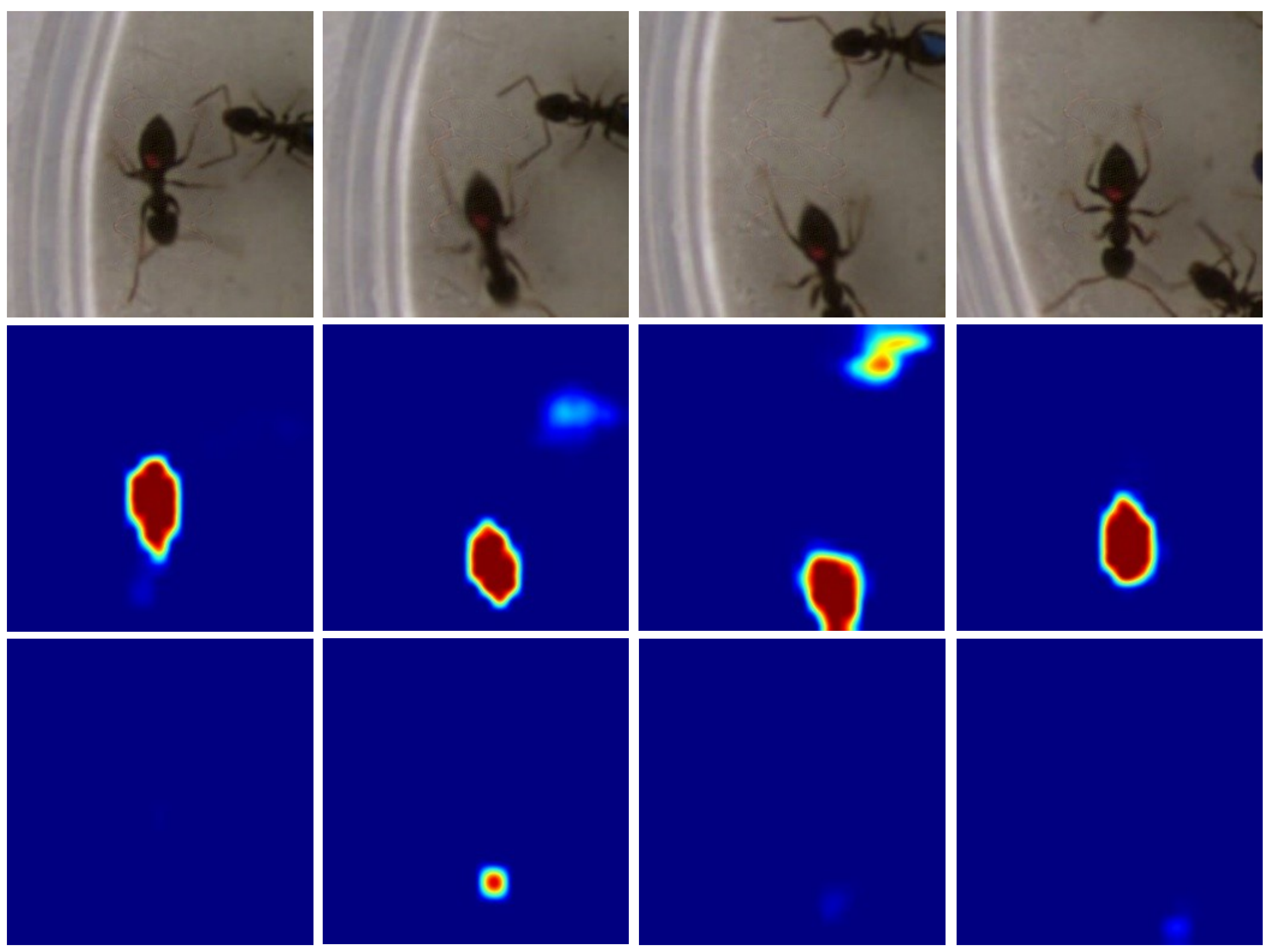}
\end{center}
\vspace{-3mm}
\caption{Search regions and their corresponding heatmaps. The first row shows search regions. 
The second row represents clean heatmaps generated by the clean template. 
The third row represents adversarial heatmaps generated by the adversarial template. }
\label{fig-heatmap}
\vspace{-3mm}
\end{figure}

\begin{figure*}[!t]
\begin{center}
\begin{tabular}{ccc}
\includegraphics[width=0.32\linewidth,height=0.35\linewidth]{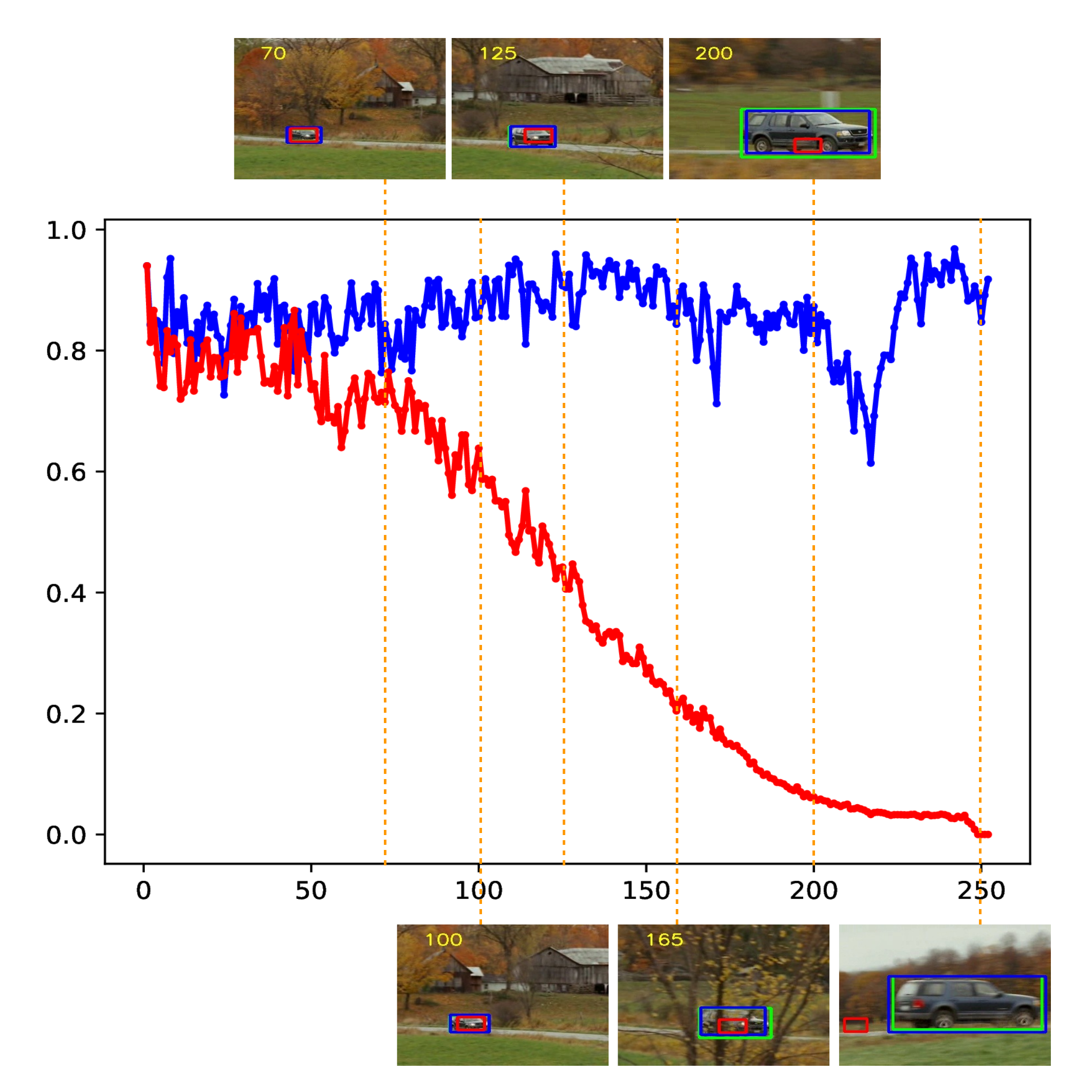} \ &
\includegraphics[width=0.32\linewidth,height=0.35\linewidth]{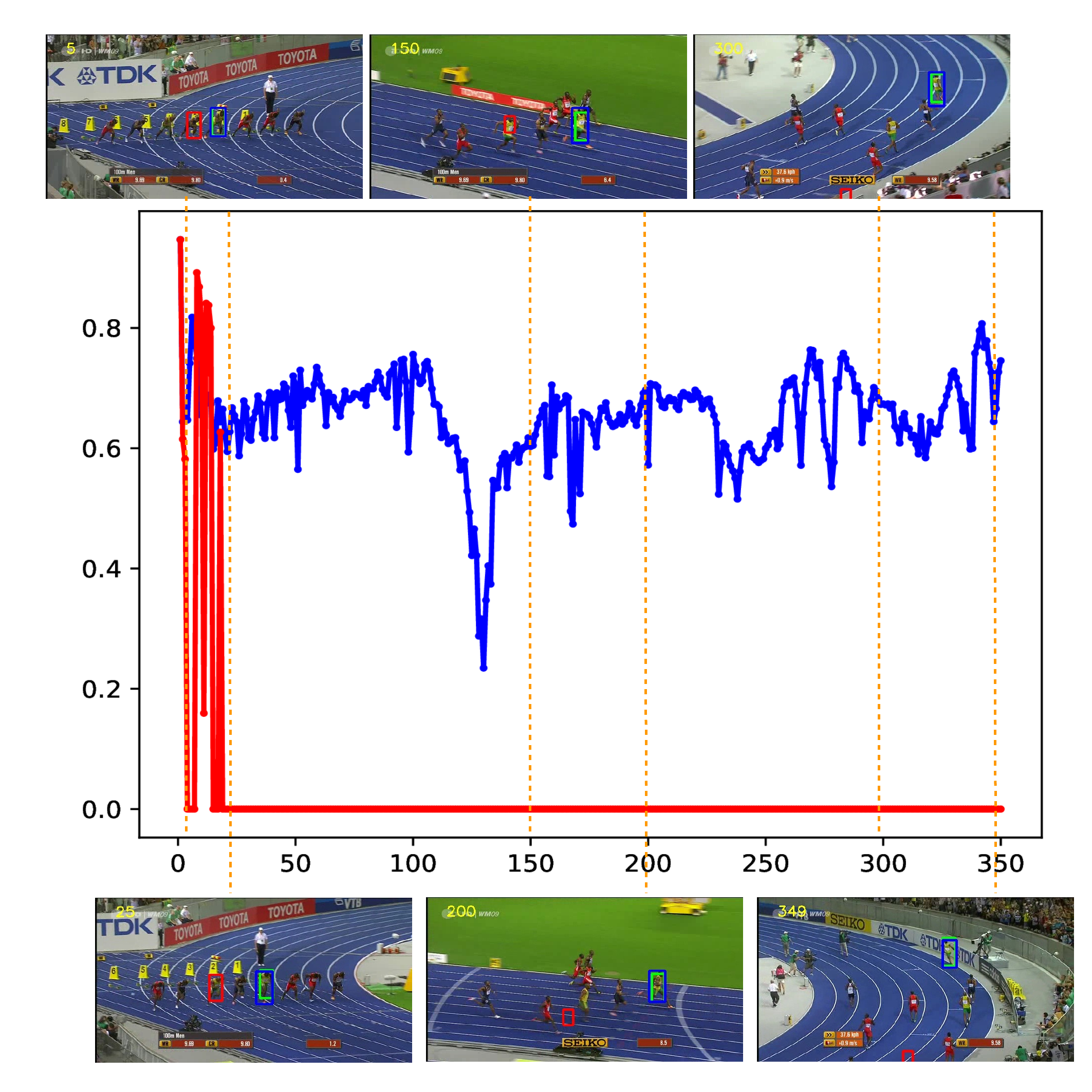} \ &
\includegraphics[width=0.32\linewidth,height=0.35\linewidth]{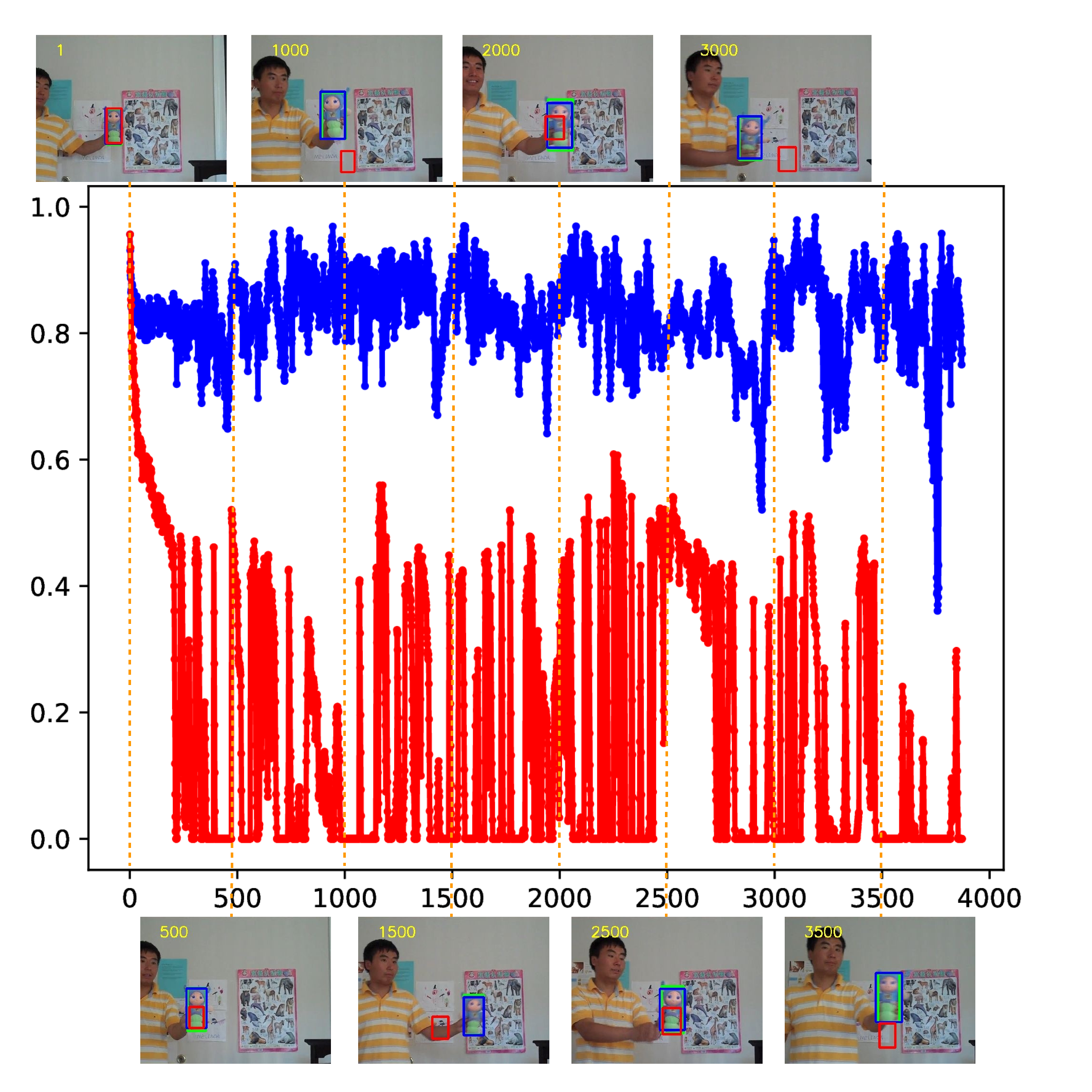}\\
(a) \footnotesize{CarScale} & (b)  \footnotesize{Bolt} & (c)  \footnotesize{Doll}\\
\end{tabular}
\end{center}
\vspace{-3mm}
\caption{Illustration of the effectiveness of generated perturbations. The green, blue and red boxes 
represent the groundtruth, original results and adversarial results, respectively. The blue and red lines 
represent the IoU's variation of the original results and adversarial results over time, respectively. 
Better viewed in color with zoom-in.}
\label{fig-otb-imgs}
\vspace{-3mm}
\end{figure*}

We propose a novel cooling-shrinking loss, composed of the cooling loss $L_C$ for interfering the heatmaps 
$M_H$ and the shrinking loss $L_S$ for interfering the regression maps $M_R$. 
To determine the location of the target, we also introduce the clean heatmaps $M_H^c$ to the computation.
In specific, the cooling loss $L_C$ is designed to cool down hot regions where the target may exist on the heatmaps, 
causing the tracker to lose the target. 
The shrinking loss $L_S$ is designed to force the predicted bounding box to shrink, leading to error accumulation 
and tracking failure. 
To compute these two losses conveniently, we reshape the clean heatmaps $M_H^c$, adversarial heatmaps
$M_H^a$, and adversarial regression maps $M_R^a$ to 2D matrices $\widetilde{M_H^c}$, 
$\widetilde{M_H^a}$, $\widetilde{M_R^a}$ with shape (N,2), (N,2), (N,4) respectively.
Then, $\widetilde{M_H^c}$ is activated with the softmax function, generating the probability of target 
$P_+$ and the probability of background $P_-$.
Based on the probability $P_+$ and a predefined threshold $\mathcal{T}$, binary attention maps $\mathcal{A}$ 
are computed, indicating locations that we are interested in.
After that, we define the cooling loss based on the difference between the confidence score of positive class $f_+$ 
and negative class $f_-$ on regions where $\mathcal{A}>0$.
We also set a margin $m_c$ in this loss to avoid any unconstrained decrease of this loss, 
leading to difficulty in controlling noises' energy. 
Similarly, we set two scale factors $R_w,R_h$ as the core of the shrinking loss and also set margins 
$m_w$ and $m_h$ as we do in the cooling loss.
The detailed mathematical formulas about the cooling-shrinking loss are shown in Algorithm~\ref{alg::CS-loss}. 
Figure~\ref{fig-heatmap} shows the effect of the cooling-shrinking loss. The second row represents the heatmaps 
produced by the clean template, and the third row represents heatmaps produced by the adversarial template. 
The adversarial heatmaps have low values on places where the target exists, making the tracker difficult to locate 
the tracked target. 
Figure~\ref{fig-otb-imgs} shows a comparison between the original results and adversarial results. 
After adding adversarial perturbations, the tracker becomes less scale-sensitive (Figure~\ref{fig-otb-imgs}(a)), less discriminative 
(Figure~\ref{fig-otb-imgs}(b)), and less target-aware (Figure~\ref{fig-otb-imgs}(c)). 
To be specific, in Figure~\ref{fig-otb-imgs}(a), the tracker produces shrinking boxes when the target actually 
grows larger, causing inaccurate results. 
In Figure~\ref{fig-otb-imgs}(b), the tracker recognizes other distractors as the target. 
In addition, in Figure~\ref{fig-otb-imgs} (c), the tracker loses the target quickly and can only re-capture it 
when it accidentally returns to the previous location.
\vspace{-3mm}
\begin{algorithm}[h]
  \caption{Cooling-Shrinking Loss}
  \label{alg::CS-loss}
  \begin{algorithmic}[1]
    \Require
      $M_H^c$: clean heatmaps;
      $M_H^a$: adversarial heatmaps;
      $M_R^a$: adversarial regression maps; 
      $\mathcal{T}$: threshold for probability;
      $m_c$: margin for classification;
      $m_w$: margin for width regression factor;
      $m_h$: margin for height regression factor;
    \Ensure
      cooling loss $L_C$;
      shrinking loss $L_S$;
    \State Reshape $M_H^c$, $M_H^a$, $M_R^a$ to 2D matrices: $\widetilde{M_H^c}$, $\widetilde{M_H^a}$, $\widetilde{M_R^a}$;
    \State $P_+,P_- = softmax(\widetilde{M_H^c})$;
    \State $\mathcal{A} = \left\{ {\begin{array}{*{20}{c}}
  1&{{P_+} \ge \mathcal{T}}\\
  0&{{P_+} < \mathcal{T}}
  \end{array}} \right.$;
    \State $f_+,f_-=\widetilde{M_H^a}*\mathcal{A}$;
    \State $R_x,R_y,R_w,R_h=\widetilde{M_R^a}*\mathcal{A}$;
    \State $L_C=\frac{1}{N}max(f_+-f_-,m_c)$;
    \State $L_S=\frac{1}{N}max(R_w,m_w)+\frac{1}{N}max(R_h,m_h)$;
  \end{algorithmic}
\end{algorithm}

\vspace{-5mm}

\vspace{-3mm}
\subsection{Implementation Details}
\vspace{-2mm}
{\flushleft \textbf{Network Architectures}:}
The perturbation-generator adopts the U-Net~\cite{u-net} architecture, which achieves superior performance 
in many pixel-level tasks. 
U-Net first downsamples feature maps multiple times and then upsamples them accordingly to make the input size 
being a power of $2$, or the output size may mismatch with the input size.
In our experiment settings, the input resolution also depends on whether to attack the template or search regions. 
Specifically, all SiamRPN-based trackers, including 
SiamRPN~\cite{SiameseRPN}, DaSiamRPN~\cite{DSiam}, SiamRPN++~\cite{SiamRPNplusplus}, 
and SiamMask~\cite{SiamMask}, adopt the same template size as $127\times127$. 
However, when working in long-term scenarios, they may use different search region sizes like 
$255\times255$ and $831\times831$, because of switching between local and global states. 
Too large size may bring a heavy computational burden, and too small size may cause the loss of 
detailed information. 
Considering all these factors, we set the input resolution of the generator as $128\times128$ 
for attacking the template and $512\times512$ for attacking the search regions. 
The gap between different resolutions is bridged by padding-cropping or bilinear interpolation. 
For example, when attacking the template, the original template with a spatial size $127\times127$ 
is first padded to $128\times128$ with zero and passed through the generator to obtain the adversarial template. 
Then, the adversarial template is cropped back to $127\times127$ again and sent into the Siamese network. 
Similarly, when attacking search regions, clean search regions with a spatial size $255\times255$ are 
first interpolated to $512\times512$ and fed into the generator to get adversarial search regions. 
Then, adversarial search regions are interpolated back to $255\times255$ again and passed to 
the Siamese network.
\vspace{-3mm}
{\flushleft \textbf{Training Dataset}:}
We use GOT-10K~\cite{GOT10K} as our training set, to cover more types of the tracked target. 
To be specific, the GOT-10K dataset includes more than 10,000 sequences and more than 500 object classes, 
showing high tracking diversity. We expect that models learned on this dataset could have better generalization 
power rather than only work on a few limited situations. 
We only use the train split of GOT-10K and uniformly sample frames with an interval of $10$ frames, and 
then crop search regions from these chosen frames. 
The template is cropped in the initial frame, and each search region is cropped based on the groundtruth 
of the last frame to simulate the situation in online tracking. 
In each training iteration, a template and $N$ search regions from the same video sequence are sent to 
our attacking model. 
In our experiments, $N$ is not larger than $15$ due to the limited GPU memory. 
\vspace{-2mm}
{\flushleft \textbf{Training Loss Function}:}
The generator is trained with the linear combination of the cooling loss, shrinking loss, and $L_2$ loss. 
The weights of these three losses can be tuned according to different biases. For example, we can increase 
the weight of the $L_2$ loss or decrease that of adversarial losses to make attack more unrecognizable. 
In our experiment setting, we choose the weights of cooling loss, shrinking loss and, $L_2$ loss as 
$0.1$, $1$, and $500$ respectively. 
The three margins $m_c$, $m_w$, $m_h$ for preventing the unconstrained decrease of adversarial 
losses are all set to $-5$.
\vspace{-3mm}
\section{Experiments}
\vspace{-2mm}
In this work, we implement our algorithm with Pytorch~\cite{Pytorch} deep learning framework. 
The hardware platform is a PC machine with an intel-i9 CPU (64GB memory) and a RTX-2080Ti GPU 
(11GB memory). 
We evaluate the proposed adversarial attack method on three datasets: OTB100~\cite{OTB2015}, 
VOT2018~\cite{VOT2018report}, and LaSOT~\cite{LaSOT}. 
To be specific, OTB100~\cite{OTB2015} contains 100 sequences, providing a fair benchmark for 
single object tracking. VOT2018~\cite{VOT2018report} is another challenging tracking benchmark, 
which simultaneously measures the tracker's accuracy and robustness. 
This benchmark includes 60 videos and ranks the trackers' performance with the 
expected average overlap (EAO) rule. 
LaSOT~\cite{LaSOT} is a recent large-scale tracking dataset, which covers 1400 videos 
with much longer time slots. 
We denote SiamRPN++ as SiamRPNpp for concise descriptions in the experiment section. 
Numerous experimental results demonstrate that our method can fool the state-of-the-art SiamRPNpp 
tracker with merely imperceptible perturbations on the search regions or template. 
We also test our perturbation-generator on another three top-performance trackers: DaSiamRPN~\cite{DSiam}, 
DaSiamRPN-UpdateNet~\cite{UpdateNet}, and DiMP~\cite{DiMP}. 
Obvious performance drop can also be observed, which shows that our method has good transferability. 
The speed of our attacking algorithm is also extremely fast. It only takes our model less than {\bf 9} ms to transform 
a clean search region to the adversarial one, and less than {\bf 3} ms to transform a clean template to the adversarial one. 
\vspace{-2mm}
\subsection{Adversarial Attack to SiamRPNpp}
{\flushleft\textbf{Attacking Search Regions Only}:}
When attacking search regions, we leave the original template unchanged and only replace clean 
search regions with the adversarial ones in each frame. 
The detailed experimental results are shown in Table~\ref{tab-attack-SR}, and 
an obvious performance drop can be observed on all three datasets.
\begin{table}[!htbp]
  \footnotesize
  \centering
  \caption{Effect of attacking search regions. 
  The third column represents SiamRPNpp's original results. 
  The fourth column represents results produced by attacking search regions. 
  The last column represents the performance drop.}
  \begin{tabular}{|c|c|c|c|c|}
  \hline
  Dataset&Metric&Original&Attack SR&Drop\\
  \hline
  \multirow{2}*{OTB100}&Success($\uparrow$)&0.696&0.349&0.347\\
  \cline{2-5}
  &Precision($\uparrow$)&0.914&0.491&0.423\\
  \hline
  \multirow{3}*{VOT2018}&Accuracy$(\uparrow)$&0.600&0.486&0.114\\
  \cline{2-5}
  &Robustness$(\downarrow)$&0.234&2.074&1.840\\
  \cline{2-5}
  &EAO$(\uparrow)$&0.414&0.073&0.341\\
  \hline
  \multirow{2}*{LaSOT}&Norm Precision$(\uparrow)$&0.569&0.219&0.350\\
  \cline{2-5}
  &Success$(\uparrow)$&0.496&0.180&0.316\\
  \hline
  \end{tabular}
  \label{tab-attack-SR}
  \vspace{-3mm}
  \end{table}
\vspace{-2mm}
{\flushleft\textbf{Attacking the Template Only}:}
When attacking the template, we only perturb the template once in the initial frame, replacing the original 
template with the adversarial one, and then leaving the rest of the tracking process undisturbed. 
The detailed experimental results are shown in Table~\ref{tab-attack-template}. 
The performance drop in this scenario is not as much as that in attacking search regions, 
because the tracker is hard to deceive by only adding minimal noises to the initial template.
\vspace{-2mm}
\begin{table}[!htbp]
  \centering
  \caption{Effect of attacking the template. 
  The third column represents SiamRPNpp's original results. 
  The fourth column represents results produced by attacking the template. 
  The last column represents the performance drop.}
  \footnotesize
  \begin{tabular}{|c|c|c|c|c|}
  \hline
  Dataset&Metric&Original&Attack T&Drop\\
  \hline
  \multirow{2}*{OTB100}&Success$(\uparrow)$&0.696&0.527&0.169\\
  \cline{2-5}
  &Precision$(\uparrow)$&0.914&0.713&0.201\\
  \hline
  \multirow{3}*{VOT2018}&Accuracy$(\uparrow)$&0.600&0.541&0.059\\
  \cline{2-5}
  &Robustness$(\downarrow)$&0.234&1.147&0.913\\
  \cline{2-5}
  &EAO$(\uparrow)$&0.414&0.123&0.291\\
  \hline
  \multirow{2}*{LaSOT}&Norm Precision$(\uparrow)$&0.569&0.448&0.121\\
  \cline{2-5}
  &Success$(\uparrow)$&0.496&0.393&0.103\\
  \hline
  \end{tabular}
  \label{tab-attack-template}
  \vspace{-2mm}
  \end{table}
\vspace{-3mm}
{\flushleft\textbf{Attacking Both Search Regions and the Template}:}
We also design a strategy that simultaneously attacks both search regions and the template. 
In this setting, we use the same generator designed for search regions to interfere with the template and search 
regions together. 
The detailed results are shown in Table~\ref{tab-attack-template-SR}. It can be seen that this strategy brings a 
slightly higher performance drop than only attacking search regions.

\begin{table}[!htbp]
\centering
\caption{Effect of attacking both search regions and the template. 
The third column represents SiamRPNpp's original results. 
The fourth column represents results produced by attacking both search regions and the template. 
The last column represents the performance drop.}
\footnotesize
\begin{tabular}{|c|c|c|c|c|}
\hline
Dataset&Metric&Original&Attack Both&Drop\\
\hline
\multirow{2}*{OTB100}&Success$(\uparrow)$&0.696&0.324&0.372\\
\cline{2-5}
&Precision$(\uparrow)$&0.914&0.471&0.443\\
\hline
\multirow{3}*{VOT2018}&Accuracy$(\uparrow)$&0.600&0.467&0.133\\
\cline{2-5}
&Robustness$(\downarrow)$&0.234&2.013&1.779\\
\cline{2-5}
&EAO$(\uparrow)$&0.414&0.073&0.341\\
\hline
\multirow{2}*{LaSOT}&Norm Precision$(\uparrow)$&0.569&0.201&0.368\\
\cline{2-5}
&Success$(\uparrow)$&0.496&0.168&0.328\\
\hline
\end{tabular}
\label{tab-attack-template-SR}
\end{table}
\vspace{-3mm}

\begin{figure}[!t]
  \begin{center}
  \begin{tabular}{cc}
  \includegraphics[width=0.46\linewidth]{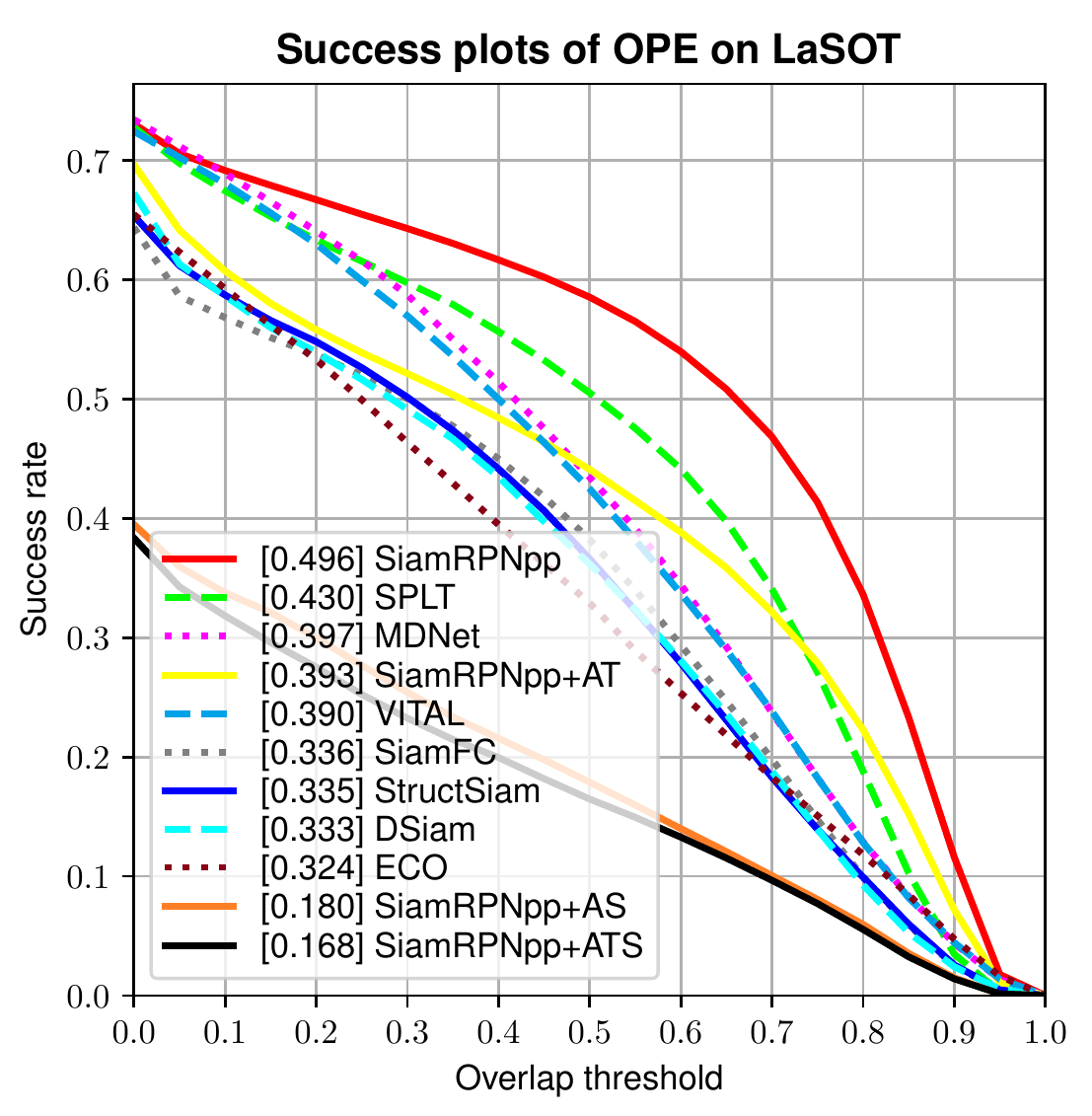} \ &
  \includegraphics[width=0.46\linewidth]{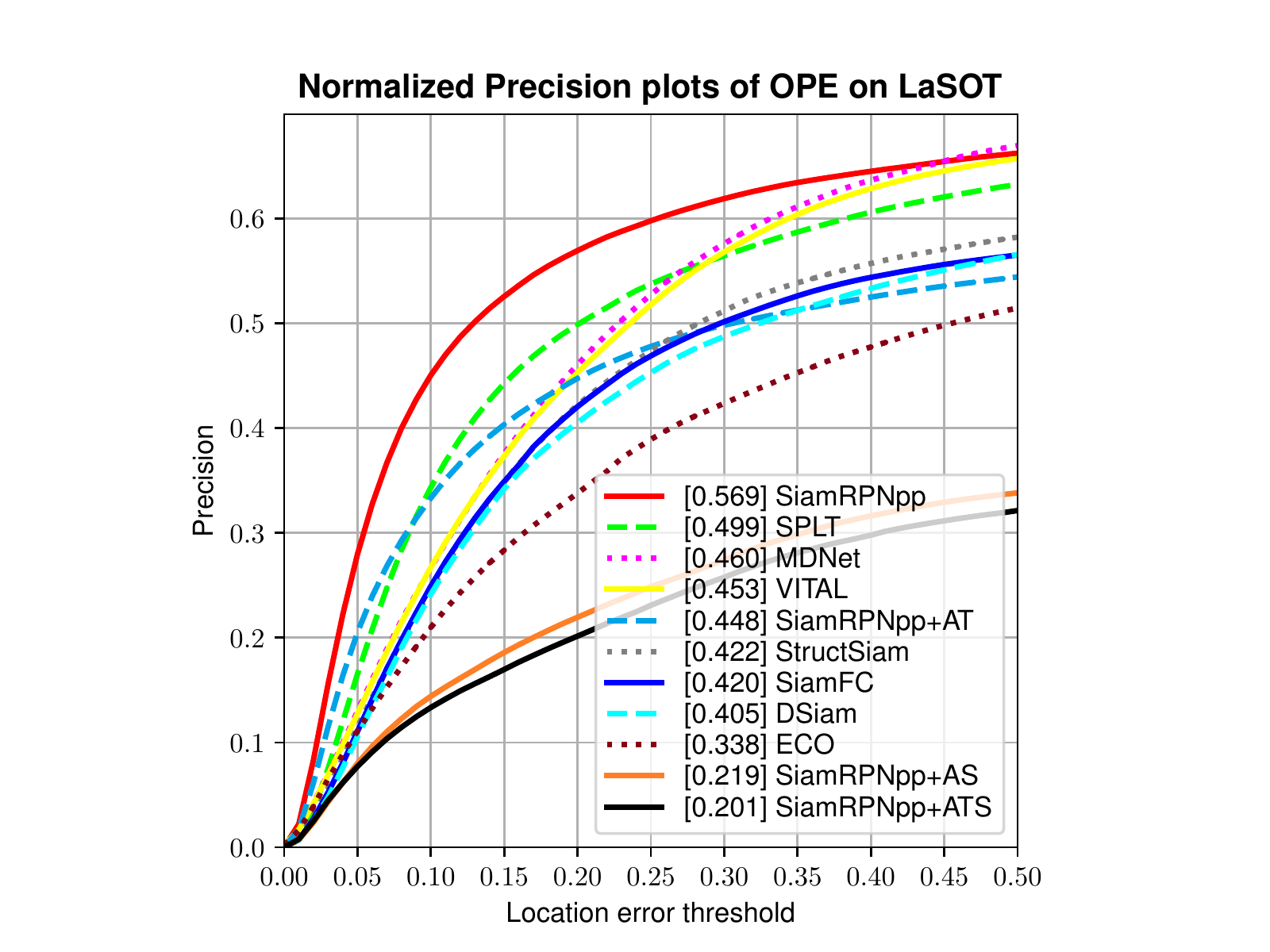}\\
  \end{tabular}
  \end{center}
  \vspace{-3mm}
  \caption{Quantitative comparison of state-of-the-art trackers on the LaSOT dataset.}
  \label{fig-siamrpnpp_lasot}
\end{figure}

\begin{figure}[!h]
\begin{center}
\includegraphics[width=0.9\linewidth,height=0.8\linewidth]{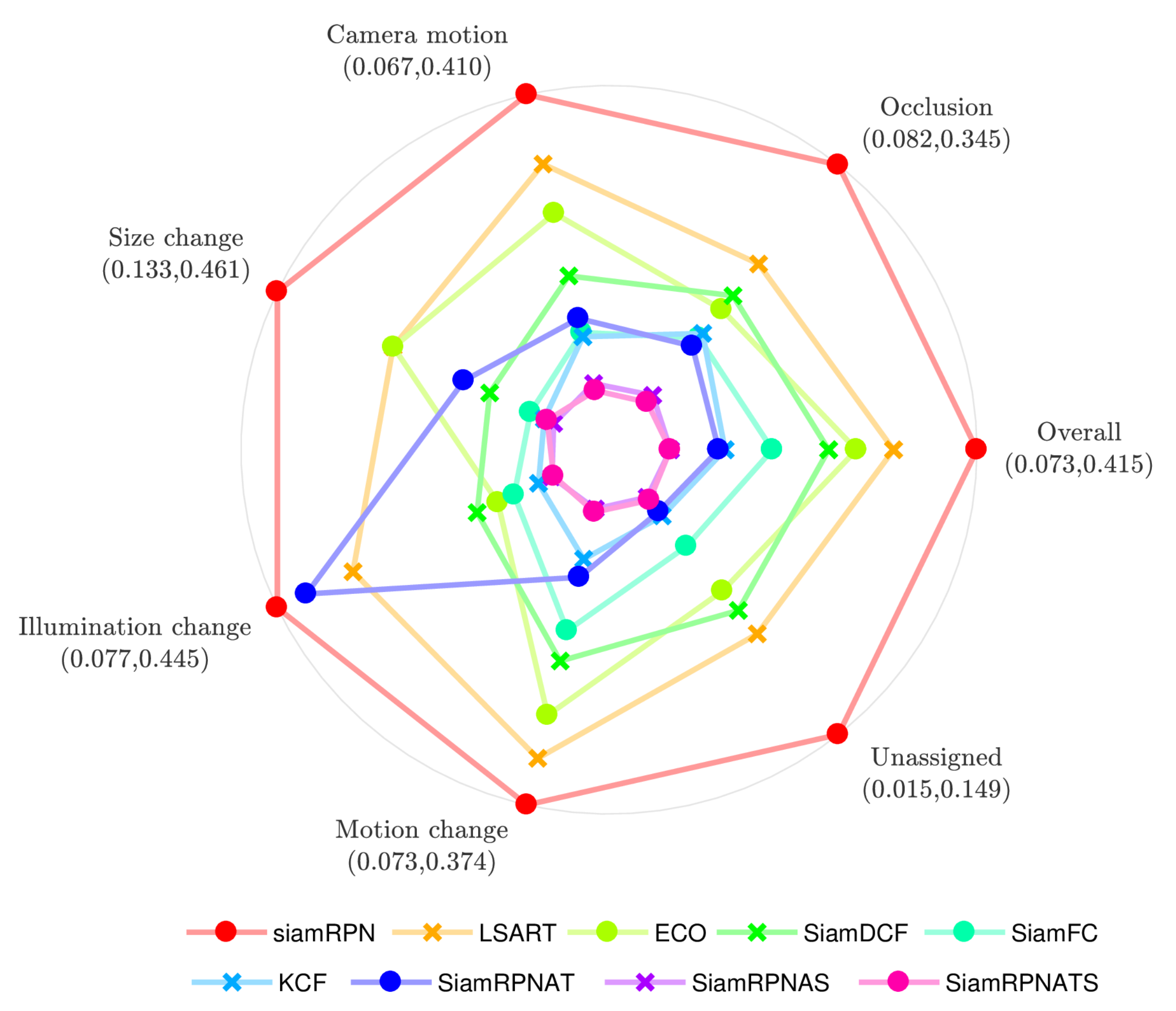}
\end{center}
\vspace{-3mm}
\caption{Quantitative analysis of different attributes on the VOT2018 dataset.}
\label{fig-attribute}
\end{figure}

We also compare the performance of SiamRPNpp~\cite{SiamRPNplusplus} and its adversarial variants:
SiamRPNpp+AT (attacking template), SiamRPNpp+AS (attacking search regions), and 
SiamRPNpp+ATS (attacking template and search regions) with other state-of-the-art trackers, such as 
MDNet~\cite{MDNet}, ECO~\cite{ECO}, SPLT~\cite{SPLT}, VITAL~\cite{VITAL}, 
StructSiam~\cite{StructSiam}, and SiamFC~\cite{SiameseFC}. 
The results are shown in Figure~\ref{fig-siamrpnpp_lasot} and Figure~\ref{fig-attribute}. 
Our adversarial attack algorithm drops the performance of SiamRPNpp significantly, making it 
obviously inferior to other top-performance trackers.

\vspace{-3mm}
\subsection{Ablation Study}
\begin{figure}[!h]
  \begin{center}
  \begin{tabular}{cc}
  \includegraphics[width=0.48\linewidth,height=0.45\linewidth]{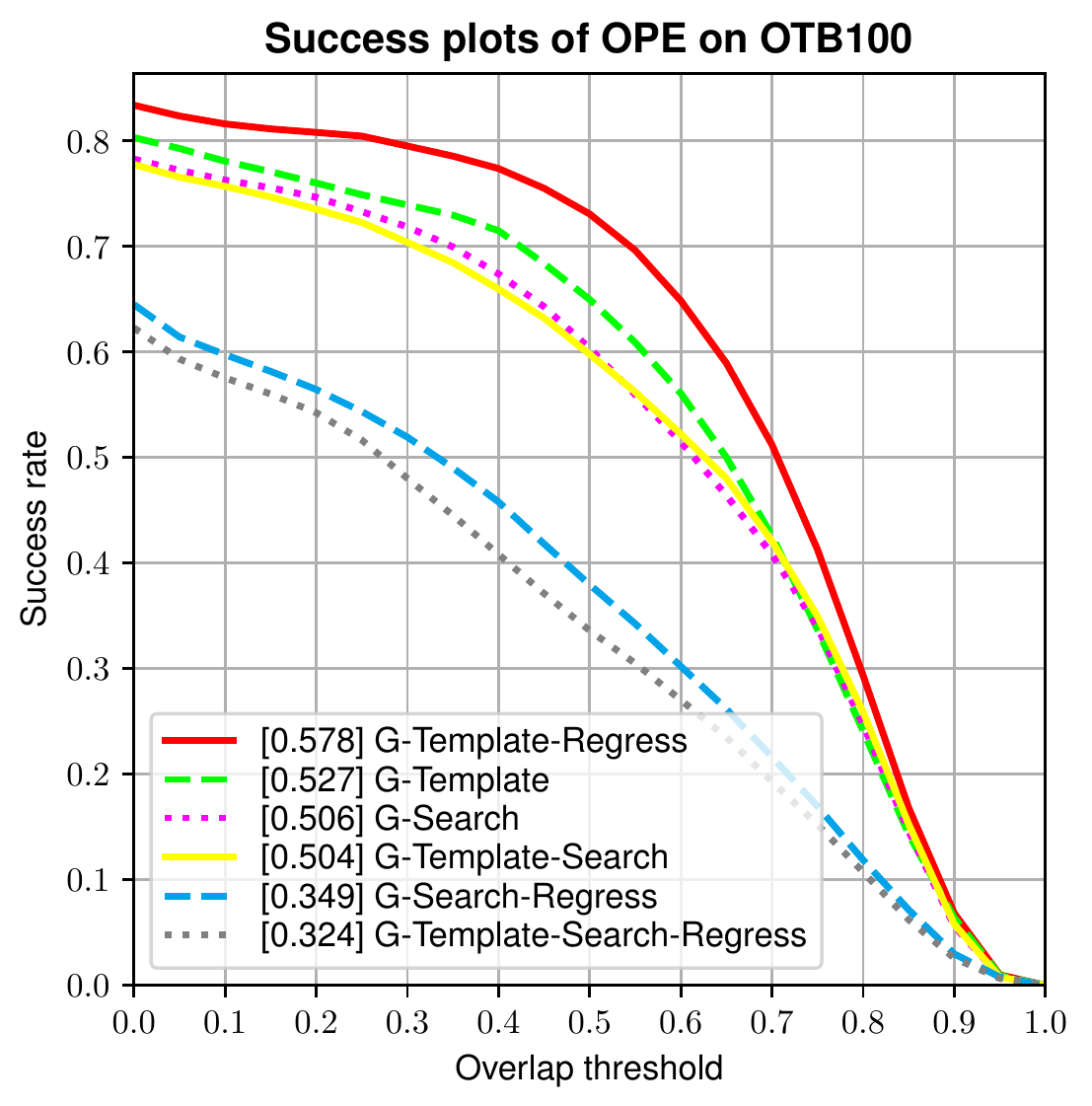} \ &
  \includegraphics[width=0.48\linewidth,height=0.45\linewidth]{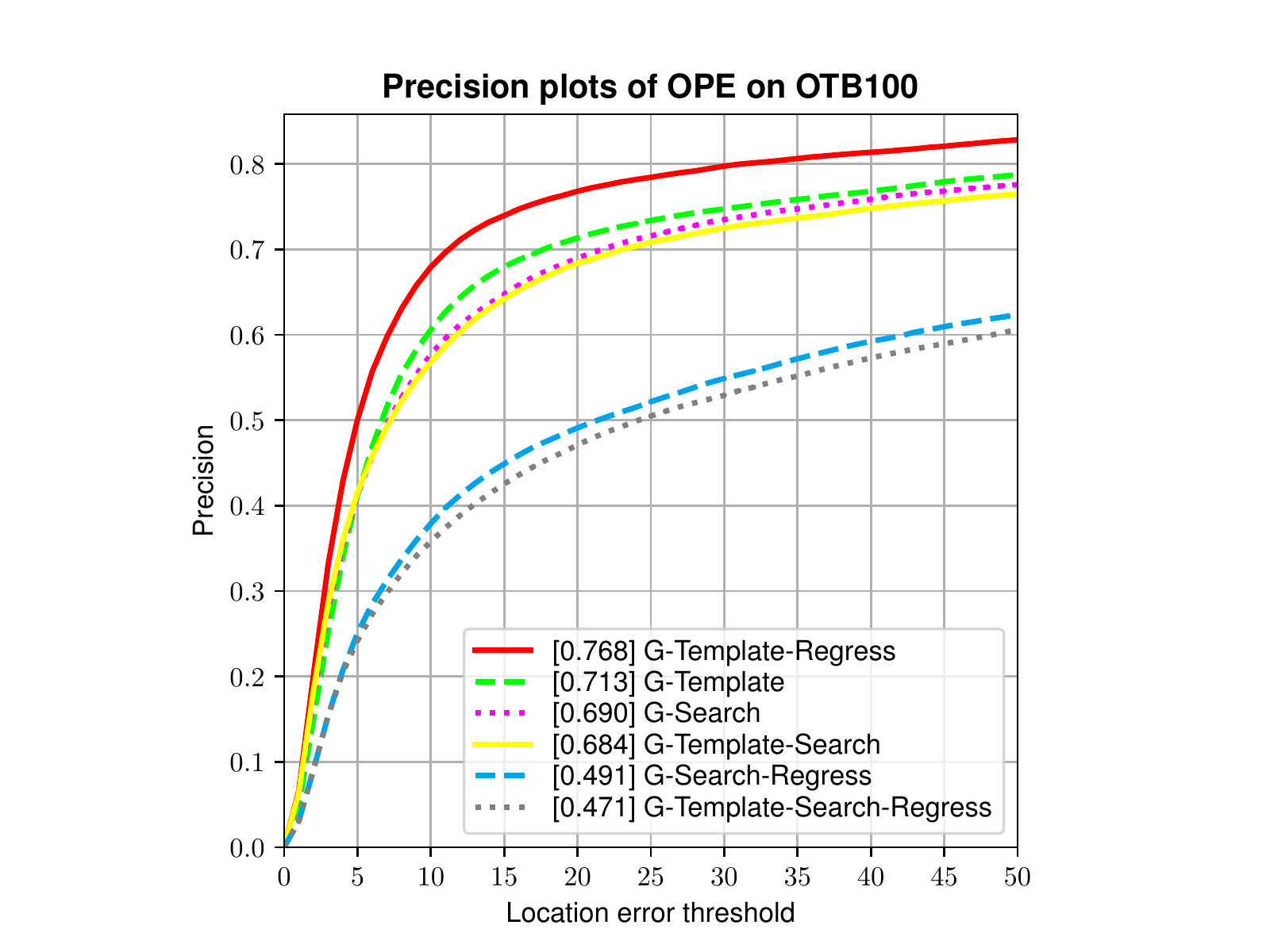}\\
  \end{tabular}
  \end{center}
  \vspace{-3mm}
  \caption{Quantitative comparisons between w/ and w/o shrinking loss on the OTB100 dataset. 
  Results with the suffix "Regress" are ones with shrinking loss.}
  \label{fig-ablation_otb}
\end{figure}

\begin{figure}[!h]
  \begin{center}
  \begin{tabular}{cc}
  \includegraphics[width=0.48\linewidth,height=0.45\linewidth]{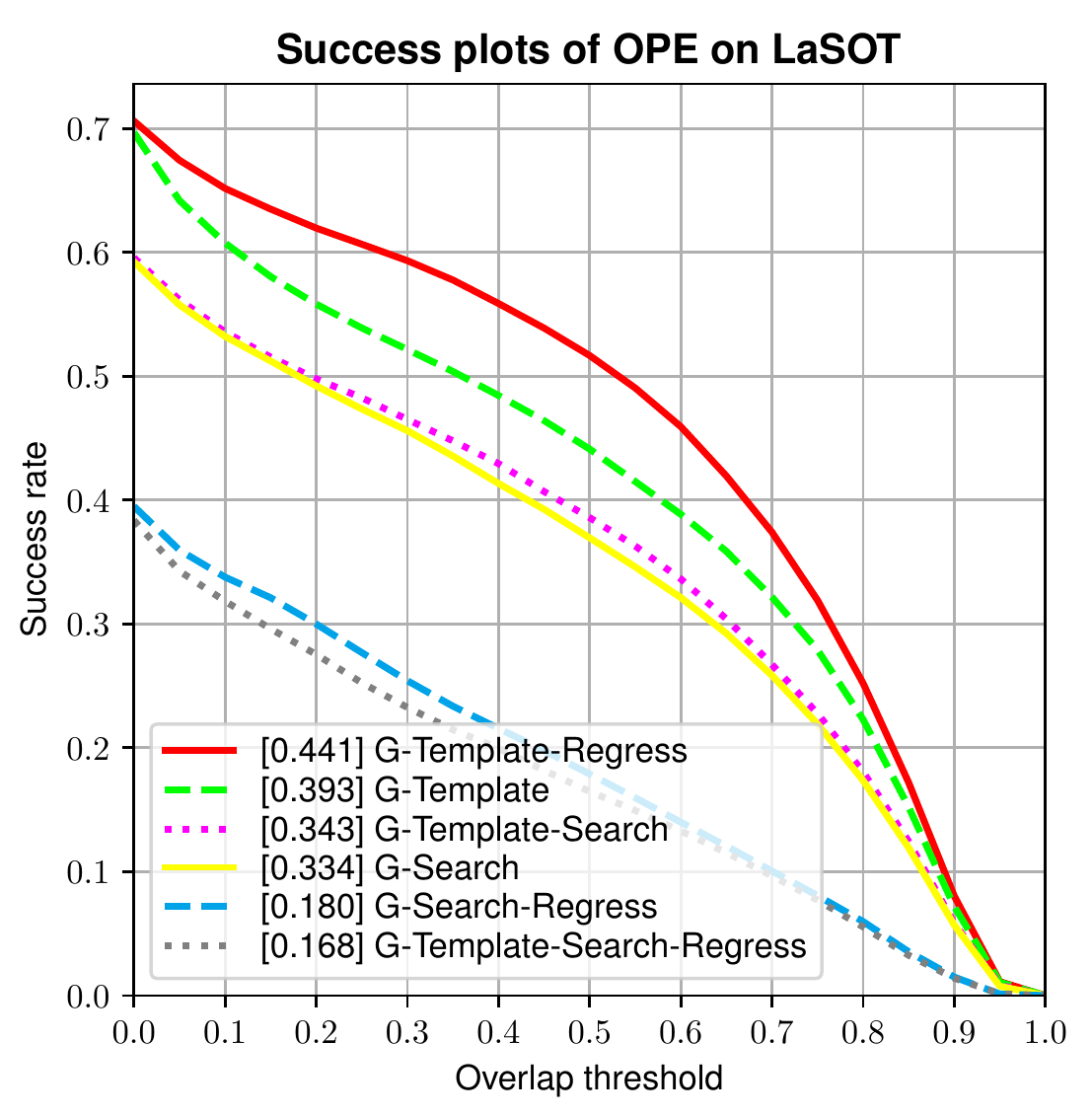} \ &
  \includegraphics[width=0.48\linewidth,height=0.45\linewidth]{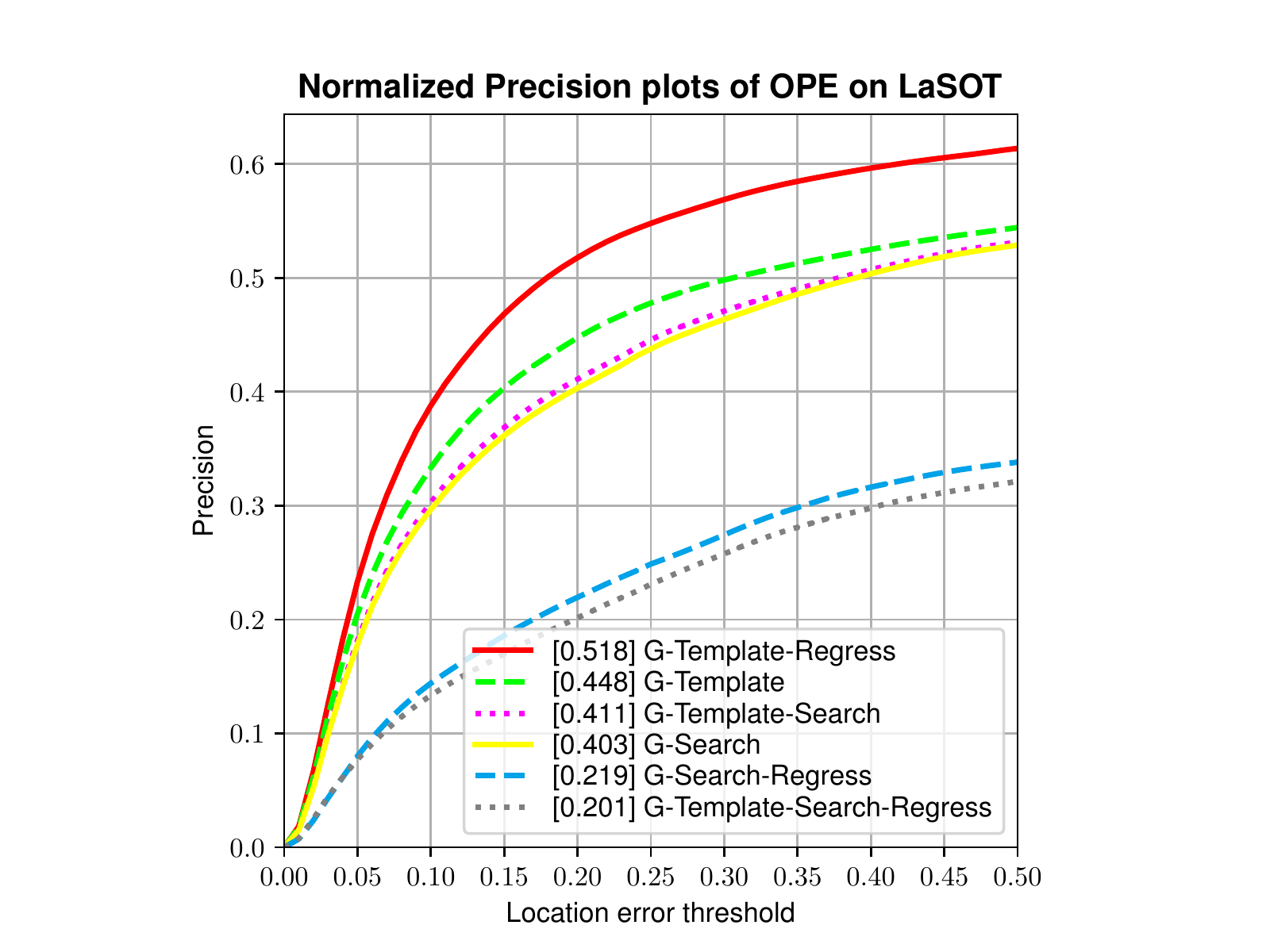}\\
  \end{tabular}
  \end{center}
  \vspace{-3mm}
  \caption{Quantitative comparisons between w/ and w/o shrinking loss on the LaSOT dataset. 
  Results with the suffix "Regress" are ones with shrinking loss.}
  \label{fig-ablation_lasot}
  \vspace{-5mm}
\end{figure}

\vspace{-2mm}
{\flushleft \textbf{Influence of Shrinking Loss}:}
We discuss the influence of the shrinking loss on the adversarial success rate in different situations. 
As explained before, the cooling loss is used to attack the classification branch, making the target invisible 
to the tracker. 
In addition, the shrinking loss is designed to disable the tracker's ability of scale estimation, thereby 
forcing the tracker to predict inaccurate bounding boxes. 
To explore the effect of the shrinking loss, we design three groups of comparison experiments on 
OTB100~\cite{OTB2015} and LaSOT~\cite{LaSOT}: 
G-Template vs. G-Template-Regress, G-Search vs. G-Search-Regress, 
and G-Template-Search vs. G-Template-Search-Regress. 
The detailed results are shown in Figure~\ref{fig-ablation_otb} and Figure~\ref{fig-ablation_lasot}. 
The shrinking loss does play a significant part when attacking search regions only and when attacking 
both search regions and template, bringing obvious extra performance drop. 
However, the shrinking loss also plays a negative part when attacking the template only, because it 
may cause the misclassification task to be suboptimal. 
To be specific, it is much more difficult to deceive the tracker by only perturbing template once than 
perturbing search regions in all frames. 
Thus, the generator cannot easily balance between cooling loss and $L_2$ loss when attacking 
only the template.
Adding an extra shrinking loss may lead to serious difficulty in training, causing worse attacking 
performance. 
In summary, the shrinking loss is helpful to attack search regions but somewhat harmful to attack 
the template.  

\vspace{-2mm}

\begin{figure}[!h]
\begin{center}
\includegraphics[width=1.0\linewidth,height=0.6\linewidth]{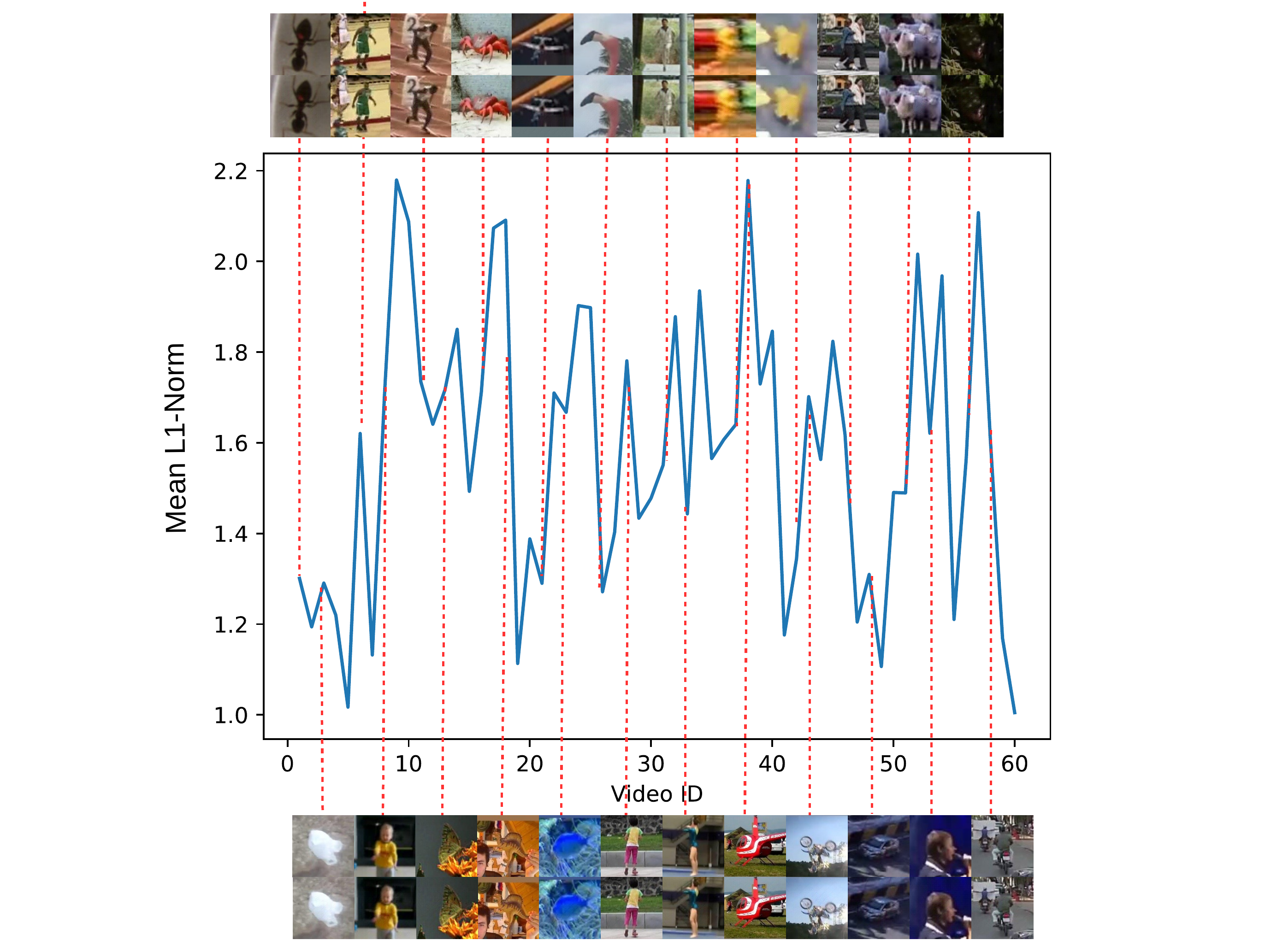}
\end{center}
\vspace{-3mm}
\caption{Visualization of clean and perturbed templates of the VOT2018 dataset. 
Better viewed in color with zoom-in.}
\label{fig-vot-template}
\vspace{-5mm}
\end{figure}

\vspace{-2mm}
{\flushleft \textbf{Influence of a Discriminator}:}
We also discuss the influence of a discriminator. Most previous neural-network-based 
adversarial attack methods~\cite{advGAN,UEA} adopt GAN structure, using a discriminator to supervise 
the adversarial output of the generator to be similar to the original input. 
However, we argue that the discriminator is not necessary. 
The reason why the $L_2$ loss and discriminator are applied is that we expect the perturbed image 
and original image to look similar. In other words, we hope that the perturbation is imperceptible. 
The $L_2$ loss can directly restrict the energy of noises and can be easily optimized. 
However, for the GAN's structure, the evolution of generator and discriminator has to be 
synchronized, which is hard to guarantee especially when the generator has many other tasks to learn. 
Thus, considering the instability of the GAN's architecture, we discard the discriminator and train 
the perturbation-generator only with the cooling-shrinking loss and $L_2$ loss.
The visualization of clean and adversarial templates from the VOT2018 dataset is shown in 
Figure~\ref{fig-vot-template}. 
Without the help of a discriminator, the perturbation generated by our method is also quite imperceptible. 

\subsection{Further Discussions}
\begin{figure}[H]
   \begin{center}
   \includegraphics[width=0.7\linewidth]{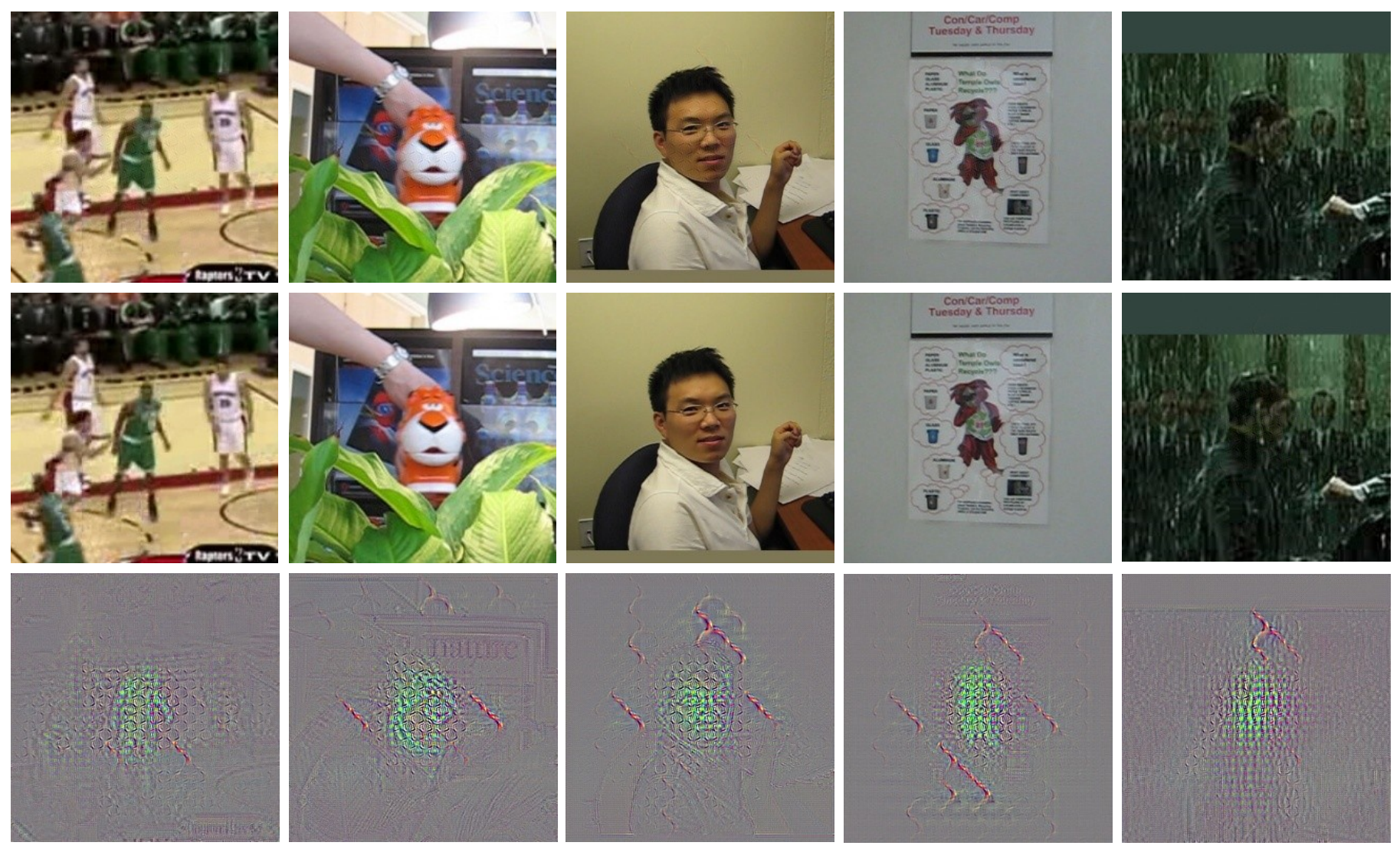}
   \end{center}
   \vspace{-3mm}
   \caption{Adversarial search regions, clean ones and their difference maps. 
   To observe pattern of difference maps clearly, the differences have been magnified by 10 times.}
   \label{fig-noise-pattern}
\end{figure}

\vspace{-2mm}
{\flushleft\textbf{Speed}:} Our method also has extremely high efficiency. When attacking search regions, our method only needs less 
than \textbf{9 ms} to process a frame, running in more than \textbf{100 FPS}. 
When attacking the template, our method needs less than \textbf{3ms} to process a whole video sequence. 
The speed of our algorithm is much faster than that of common video flow and that of most real-time trackers, 
indicating that it is also imperceptible in terms of time consumption.
\vspace{-3mm}
{\flushleft\textbf{Noise Pattern}:}
The adversarial search regions, the clean ones and their difference maps are shown in Figure~\ref{fig-noise-pattern}. 
It can be seen that the perturbation mainly focuses on the tracked target, leaving other regions almost not perturbed.
\vspace{-5mm}
{\flushleft\textbf{Comparasion with Other Noises}:} As shown in Figure~\ref{fig-noise-compare} and Table~\ref{tab-noise-compare}, compared with impulse noises 
and gauss noises, our adversarial perturbation is far more imperceptible but causes much larger performance drop.
\vspace{-3mm}
\begin{figure}[H]
   \centering
   \includegraphics[width=.8\linewidth]{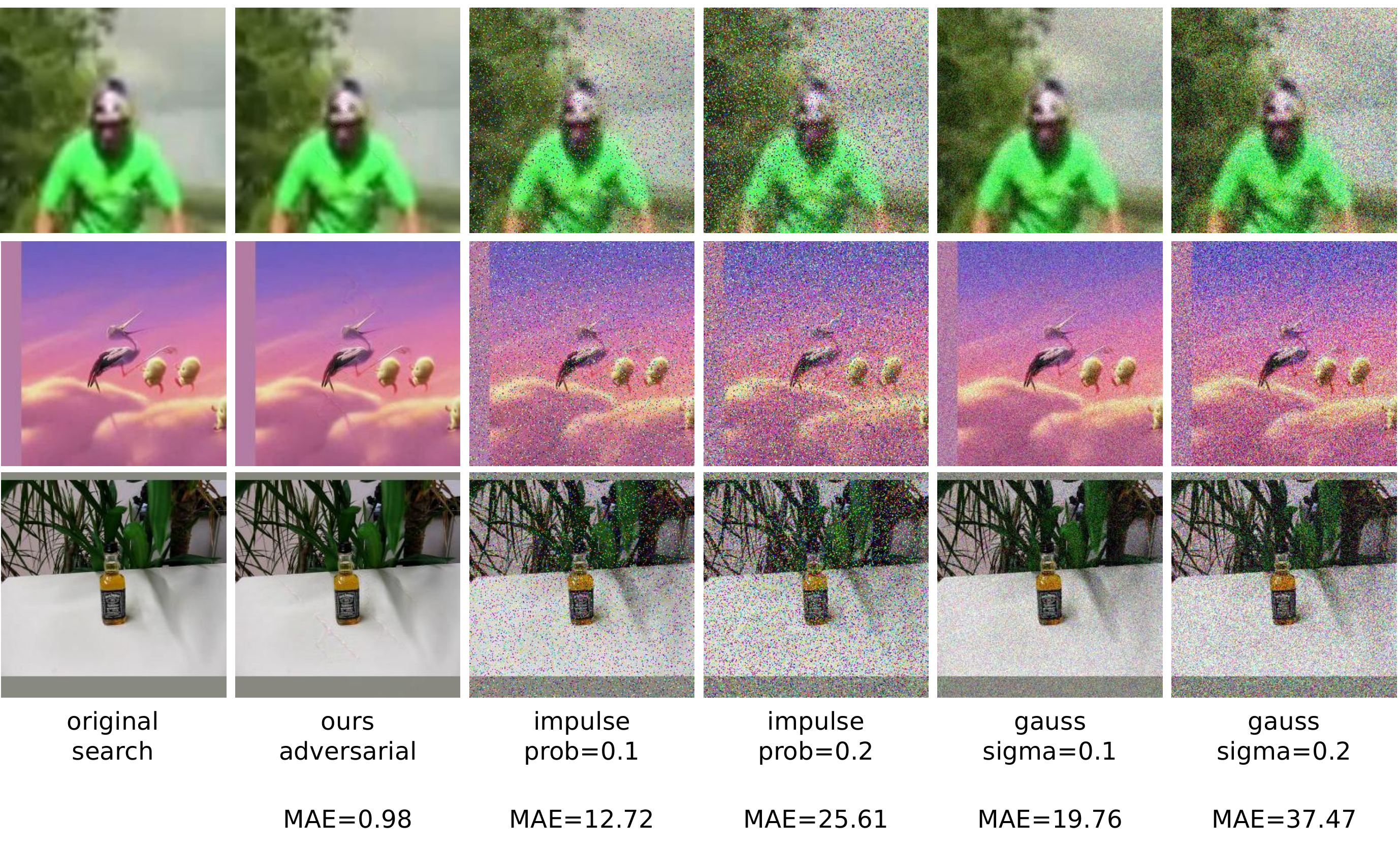}
   \vspace{-2.5mm}
   \caption{Search regions with different kinds of noises. MAE represents mean absolute error.} \label{fig-noise-compare}
   \label{fig-noise-compare} 
  \end{figure} 
\vspace{-5mm}
\begin{table}[!htbp]
   \centering
   \caption{Comparison with other kinds of noises.}
   \resizebox{\columnwidth}{!}{
   \begin{tabular}{|c|c|c|c|c|c|c|c|}
   \hline
   Dataset&Metric&original&ours&impulse 0.1&impulse 0.2&gauss 0.1&gauss 0.2\\
   \hline
   \multirow{2}*{OTB100}&Success$(\uparrow)$&0.696&0.349&0.486&0.389&0.553&0.389\\
   \cline{2-8}
   &Precision$(\uparrow)$&0.914&0.491&0.656&0.535&0.727&0.542\\
   \hline
   \multirow{1}*{VOT2018}&EAO$(\uparrow)$&0.414&0.073&0.117&0.084&0.170&0.108\\
   \cline{2-8}
   \hline
   \end{tabular}
   }
   \label{tab-noise-compare}
   \end{table}
   \vspace{-5mm}
{\flushleft\textbf{Transferability}:} All aforementioned experiments are based on the SiamRPNpp~\cite{SiamRPNplusplus} tracker. 
   To test our method's transferability, we also apply our trained perturbation-generator 
   to another three state-of-the-art trackers: DaSiamRPN~\cite{DSiam}, 
   DaSiamRPN-UpdateNet~\cite{UpdateNet}, and DiMP~\cite{DiMP}. 
   Although all these trackers have templates and search regions, they are quite different from SiamRPNpp. 
   To be specific, compared with SiamRPNpp, DaSiamRPN adopts a simpler backbone. 
   DaSiamRPN-UpdateNet proposes a learnable way to update the template, further improving DaSiamRPN's performance. 
   DiMP uses an online discriminative model to roughly determine the location of the target on the search region, 
   and then applies a state estimation module to predict precise bounding boxes. 
   We use these three trackers as the baseline algorithms and then add perturbations to their search region in each frame.
   We make experiments on the LaSOT~\cite{LaSOT} dataset.
   The detailed results about attacking DaSiamRPN, DaSiamRPN-UpdateNet, DiMP are shown in Table~\ref{tab-transfer}. 
   Although our attacking algorithm is designed for and trained with SiamRPNpp, this method can 
   also effectively deceive other state-of-the-art trackers, causing obvious performance drop, 
   which demonstrates the transferability of our attacking method.
   \vspace{-1mm}
   \begin{table}[!htbp]
   \footnotesize
   \centering
   \caption{Adversarial effect on other state-of-the-art trackers. 
   From top to bottom, three trackers are DaSiamRPN, DaSiamRPN+UpdateNet and DiMP.}
   \begin{tabular}{|c|c|c|c|}
   \hline
   Tracker& &Success$(\uparrow)$&Norm Precision$(\uparrow)$\\
   \hline
   \multirow{3}*{DaSiamRPN}&Original&0.458&0.544\\
   \cline{2-4}
   &Adversarial&0.400&0.479\\
   \cline{2-4}
   &Drop&0.058&0.065\\
   \hline
   \multirow{3}*{UpdateNet}&Original&0.465&0.549\\
   \cline{2-4}
   &Adversarial&0.399&0.478\\
   \cline{2-4}
   &Drop&0.066&0.071\\
   \hline
   \multirow{3}*{DiMP50}&Original&0.559&0.642\\
   \cline{2-4}
   &Adversarial&0.492&0.567\\
   \cline{2-4}
   &Drop&0.067&0.075\\
   \hline
   \end{tabular}
   \label{tab-transfer}
   \end{table}

\vspace{-5mm}
\section{Conclusion}
In this study, we present an effective and efficient adversarial attacking algorithm for deceiving single 
object trackers. A novel cooling-shrinking loss is proposed to train the perturbation-generator. 
The generator trained with this adversarial loss and $L_2$ loss can deceive SiamRPN++ at a high 
success rate with imperceptible noises. 
We show that a discriminator is not necessary in adversarial attack of the tracker because 
the combination of the adversarial loss and $L_2$ loss has already achieved our goal. 
Besides SiamRPN++, our attacking method has impressive transferability and effectively attacks 
other recent trackers such as DaSiamRPN, DaSiamRPN-UpdateNet, and DiMP. 
Our algorithm is also quite efficient and can transform clean templates/search regions to adversarial 
ones in a short time interval.

\vspace{-2mm}
\small{{\flushleft \textbf{Acknowledgement.}} The paper is supported in part by the National Key R$\&$D 
Program of China under Grant No. 2018AAA0102001 and National Natural Science Foundation of China 
under grant No. 61725202, U1903215, 61829102, 91538201, 61751212 and the Fundamental Research 
Funds for the Central Universities under Grant No. DUT19GJ201.}

{\small
\bibliographystyle{ieee_fullname}
\bibliography{egbib}
}

\end{document}